# An unsupervised approach for semantic place annotation of trajectories based on the prior probability

Junyi Cheng [a], Xianfeng Zhang [a]*, Peng Luo [b], Jie Huang [a], Jianfeng Huang [a]

[a] Institute of Remote Sensing and Geographic Information Systems, Peking University, 5 Summer Palace Road, Beijing, 100871, China;
[b] Chair of Cartography, Technical University of Munich, 80333 Munich, Germany.

**Abstract:** Semantic place annotation can provide individual semantics, which can be of great help in the field of trajectory data mining. Most existing methods rely on annotated or external data and require retraining following a change of region, thus preventing their large-scale applications. Herein, we propose an unsupervised method denoted as UPAPP for the semantic place annotation of trajectories using spatiotemporal information. The Bayesian Criterion is specifically employed to decompose the spatiotemporal probability of the candidate place into spatial probability, duration probability, and visiting time probability. Spatial information in ROI and POI data is subsequently adopted to calculate the spatial probability. In terms of the temporal probabilities, the Term Frequency–Inverse Document Frequency weighting algorithm is used to count the potential visits to different place types in the trajectories, and generates the prior probabilities of the visiting time and duration. The spatiotemporal probability of the candidate place is then combined with the importance of the place category to annotate the visited places. Validation with a trajectory dataset collected by 709 volunteers in Beijing showed that our method achieved an overall and average accuracy of 0.712 and 0.720, respectively, indicating that the visited places can be annotated accurately without any external data.

**Keywords:** Trajectory data mining; Semantic annotation; Place matching; Location-aware computing; Spatiotemporal probability

# 1. Introduction

Semantic place annotation refers to the process of determining the most likely visited places hidden in the raw trajectory data by contextual information (e.g., spatial and temporal information). With the increasing popularity of mobile terminals and the development of global positioning technology, massive amounts of human movement trajectory data have been collected by mobile terminals. Many studies have investigated various spatiotemporal patterns in the trajectory data [1-3] whereby a spatiotemporal trajectory is typically expressed as a series of coordinates, clusters or geographic regions. A trajectory records the purpose and habits of individuals' travel patterns. However, the absence of semantic information related to the visited places prevents the comprehensive understanding of the daily behavior and complicates its subsequent usage. Understanding the semantics contained in the massive spatiotemporal trajectory data is of great significance for personalized service recommendations and predictions, the advance prediction and prevention of emergencies, and personal health monitoring. Therefore, much attention has been focused on combining geographic and temporal information for the automated semantic annotation of the trajectories before data mining [4-7].

Existing methods for semantic place annotation typically include two steps, trajectory segmentation and semantic annotation. The process of subdividing the original trajectory into sequences in which "moving" and "stop" episodes alternately appear is called trajectory segmentation [8]. In particular, a "stop" is not completely stationary, rather it denotes staying in a relatively small neighborhood for an extended period. It is assumed that if an individual has stopped, they must be doing some particular activity or waiting at some specific places, and therefore the stop itself is meaningful [9]. Semantic annotation is also a semantic enrichment process applied by integrating different types of data to obtain meaningful information from the analysis of a raw trajectory. The annotation of the moving episode typically includes inferring the travel mode, while the annotation of the stop episode determines the most probable places visited by the individual in their stops. Compared with moving episodes, stops exhibit richer semantic information and are considered more important in trajectory data mining.

However, the accuracy of place annotation is limited by the following factors: 1) the positioning terminal often has a low accuracy and cannot effectively match the real location, 2) place ambiguity is a common occurrence, particularly for multiple place types in the same area (e.g., multiple shops or restaurants in a shopping mall), and 3) temporal ambiguity is also observed, for example, visiting the same place for different purposes results in multiple temporal patterns.

Many studies have recently been conducted on the semantic place annotation of the trajectories. These approaches have usually tried to determine the most likely visited places hidden in the trajectory data by multi-source information. However, these studies either ignore the temporal information of the stop or rely on annotated samples or other external data [5, 7]. Moreover, it is usually very tough to obtain large-scale annotated data. Therefore, a semantic annotation model that can effectively combine temporal and spatial information without supplementary data is urgently required.

In this study, we develop a prior probability-based method for unsupervised semantic annotation of trajectory stop places. The contribution of our work is as follows:

1) A prior probability statistical method for semantic trajectory annotation that does not require annotated data or other external data is proposed. With only the trajectory data, our method can extract the visiting time and duration patterns for the visited places of different types. It begins with the trajectory dataset itself and is not restricted by regions and people; thus, our method solves the dependence on annotated data in the existing methods and can greatly improve the availability of trajectory data.

2) A probability model that considers the spatial characteristics, visiting time and duration of the stop episode fully is created to address the issue of place ambiguity. Our model is able to use the temporal characteristics and spatial location comprehensively to calculate the spatiotemporal probability of different places, combining the spatiotemporal probability of the place and the importance of the place category to annotate the visited place.

3) A method that determines the spatial probability by combining ROI (region of interest) and POI (point of interest) data is proposed to increase the accuracy of the spatial probability calculations, fully employing the topological characteristics provided by the ROI data and the spatial distribution characteristics of the POI.

## 2. Related work

### 2.1 Trajectory segmentation

Current trajectory segmentation methods can be classified into supervised, unsupervised, and semi-supervised approaches [10, 11]. The supervised method uses labeled samples for training and learns the knowledge to generate sub-trajectories (e.g. the recently proposed WS-II) [12]. The semi-supervised approach mainly includes RGRASP-SemTS [13] and CoExDBSCAN [14], which uses a combination of user-labeled meaningful segments and unlabeled data to segment the trajectory. Unfortunately, most trajectory datasets do not contain labels, and it is challenging to label big trajectory datasets manually. Consequently, the unsupervised method is an important topic in trajectory segmentation.

Unsupervised segmentation algorithms can be divided into four categories, namely, rule-based, clustering-based, sliding-window-based, and cost-function-based approaches. Rule algorithm detects the trajectory sequence that meets the required spatiotemporal conditions via a threshold. Commonly used thresholds include spatial distance [15], time length [16], speed [17], and angle change thresholds [18]. Although a rule-based method is easy to implement and relatively intuitive, it is not robust to noise. Cost-function-based approach partitions a trajectory to build the most homogeneous segments by minimizing a cost-function, such as GRASP-UTS and TS-MF based on the minimum description length principle [10, 19]. A sliding-window-based approach determines where the moving object changed its behavior within a fixed-size sliding window (kernel) by deriving the local features or interpolation (e.g., OWS [20] and SWS [11]). While these two types of methods can effectively detect changes in the motion state, they are not suitable for the semantic mining of the visited places. This is because an individual is not necessarily completely static when visiting a place such as walking in a park.

A major application of clustering-based segmentation is the detection of stop-and-move patterns [21], which describe the behavior of an object that stays within a region for a period of time. Clustering-based methods detect stops by identifying continuous sequences adjacent in space and time. Moreover, these methods redefine the concepts of neighborhood and density to find spatiotemporal nearby points. CB-SMOT is a key clustering-based algorithm that redefines the neighborhood concept in DBSCAN, taking the longest continuous sequence with an average speed less than the speed threshold as the neighborhood [22]. If the continuous sequence time exceeds the time threshold, it is considered to be a stop, otherwise it is considered to be a moving episode. However, the methods may introduce extra moving points at the beginning and end of the extracted point sequence. To overcome this weakness, TrajDBSCAN is proposed and assumes that the distances to the core point from all points in the neighborhood should be less than the distance threshold [23]. In addition, the time length of the sequence is used to replace the density concept in DBSCAN. This allows for the application of TrajDBSCAN to trajectories of different sampling intervals. Subsequent work has been developed using this clustering-based concept [24-26]. Such methods are data-driven and relatively robust against noise [27]. Therefore, in the current paper, a clustering-based method is adopted for the extraction of the stop episodes and to segment the trajectory data.

## 2.2 Semantic annotation

The semantic enrichment process annotates various semantic information in trajectories fused with different sources of information. Hence, different semantic trajectory models with semantic web standards have been proposed for representing trajectory episodes and contextual information, such as STEP [28], FrameSTEP [29], Master [30], SEMANTIC-SEG [7] and SEPSIM [31]. This method can apply reliable external resources (e.g. a geographic knowledge base) and movement features to label the trajectories in question. However, such studies are limited to a basic reasoning mechanism using spatiotemporal concepts, properties, or relationships [31]. Due to temporal and spatial ambiguity, it is challenging to annotate visited places hidden in the trajectory. In the context of trajectory data mining, this work positions itself in the reasoning stage.

In order to solve place ambiguity, the features extracted from multi-source data should be comprehensively utilized to analyze and infer the places most likely to be visited. Commonly used features can be divided into geographic-, temporal-, individual-related features. Features related to geographic data include spatial topological relationships [9, 32], distance [33, 34], the distribution of place types [4, 6]. Time-related features typically include the start time and duration of stop episodes [7, 35] and activity history related features, such as accumulated duration and frequency. Common individual-related features are the individual's occupation [36], age [37] and home address [38]. In addition to these features, other information is also selected, such as place popularity and moving patterns typically obtained from social media and review sites [7, 36, 39]. Individual-related features are challenging to obtain and are guarded by privacy restrictions, and consequently they are rarely employed in large-scale applications. Furthermore, the distribution of samples obtained from social media is often biased, and retraining is required following a change of the region. Such information is also relatively difficult to acquire. Geographic and temporal information, which can be obtained easily and is directly related to the travel patterns of individuals, is widely used in previous studies.

On the other hand, current methods for reasoning the visited places can be grouped into rule-based, machine learning-based, and probability-based approaches. The rule-based method annotates places by formulating temporal and spatial rules [40, 41]. For example, the most commonly used spatial rule takes the nearest POI as the visited place [42]. However, the rules are relatively arbitrary, preventing the place and temporal ambiguity from being solved effectively. The machine learning-based method uses labeled data for supervised learning and employs the trained model to annotate the visited places. Commonly used machine learning models in semantic annotation include decision trees [41, 43], random forests [34, 35], and neural networks [44]. A recent study has established a random forest model based on temporal, spatial and sequential features for automatically estimating the semantic meanings of personal locations [6]. This method is able to employ a variety of information and is suitable for large-scale datasets, yet it relies on labeled data for training [38, 44-46], and retraining is required after the region is changed. The probability-based method calculates the probability of each visiting place based on the spatiotemporal features. The majority of studies using the probability-based method employ spatial probability. For example, Yan established a spatial probability model based on a two-dimensional Gaussian distribution function [4]. Many researchers subsequently followed a similar approach to this spatial probability model [5, 7]. Unlike the aforementioned methods, some recent studies have adopted ROI data to establish a spatial probability model based on spatial topological relationships [9]. Current studies use either the ROI or POI when establishing the spatial probability model, while research work that comprehensively integrates information from different types of geographic data is still lacking. For the formulation of the temporal probability, previous studies generally calculate the probability directly from the activity log or employ social media to extract the required patterns. For example, Gong calculated the visiting time probability for different place types from the activity log [5]. Gao adopted review data to determine the visiting time and duration probabilities of different place types [7]. It can be seen that the aforementioned methods either ignore the temporal information or rely on external data. Therefore, a probability model that fully makes use of spatiotemporal characteristics without using any external data is proposed in this paper. Table 1 summarizes the difference of our method and the existing approaches.

Table 1. A summary of the difference between our method and the existing approaches.

| Method | Use spatial relationship? | Use temporal information? | Use place category distribution? | Require training? | Require external data? |
|---|---|---|---|---|---|
| Graaff (2016) [47], Noureddine (2020) [48] | **yes** | no | no | **no** | **no** |
| Gong (2016) [5] | **yes** | **yes** | no | **no** | yes |
| Lv (2016) [6] | no | **yes** | **yes** | yes | **no** |
| Zhang (2018) [46] | **yes** | **yes** | no | yes | **no** |
| Zhang (2019) [49] | **yes** | no | no | **no** | **no** |
| Bermingham (2019) [9] | **yes** | no | **yes** | **no** | **no** |
| Gao (2020) [7] | **yes** | **yes** | no | **no** | yes |
| Our method | **yes** | **yes** | **yes** | **no** | **no** |

## 3. Methodology

### 3.1 Overview of the framework

The proposed semantic place annotation framework is illustrated in Fig.1, which includes trajectory collection, stop and candidate place extraction, spatiotemporal probability calculation of candidate places, and semantic annotation of visited places. Throughout this work, the Unsupervised Place Annotation with Prior Probability (UPAPP) method is developed. First, the stops are extracted to segment a trajectory, and the relevant attributes of the stops are then calculated. Following this, a spatiotemporal probability model for the candidate places is created, which is decomposed into the spatial, duration and visiting time probabilities. Both POI and ROI data are used to establish a spatial probability calculation method. For the duration and visiting time probabilities, we propose an unsupervised prior probability statistical method that does not require supplementary data. The prior probabilities of different place types are extracted based on the potential visits of the trajectory. Once the spatiotemporal probabilities of the candidate places are obtained, the probability of visiting a place is calculated by the combination with the place type importance for the subsequent semantic annotation.

In the following, we provide some key definitions that are used throughout this paper.

**Definition 1**: Stop $SP = (x, y, t_{start}, dur)$ denotes a stop episode and consists of a series of continuous spatiotemporal trajectory points, where $(x, y)$ is the coordinate of the stop center, which is calculated by the average coordinate of all spatiotemporal points in that stop; $t_{start}$ represents the start time or visiting time of the stop; and *dur* indicates the time difference between the end and the start of the stop.

**Definition 2**: Stop region $R_{SP} = (x, y, r)$ indicates the spatial range covered by the circle whose radius represents the spatial uncertainty and noise present in a real trajectory stop. In addition, the stop center is chosen as the center of the circle, and the stop radius equals to the maximum distance to the stop center from all spatiotemporal points in the stop.

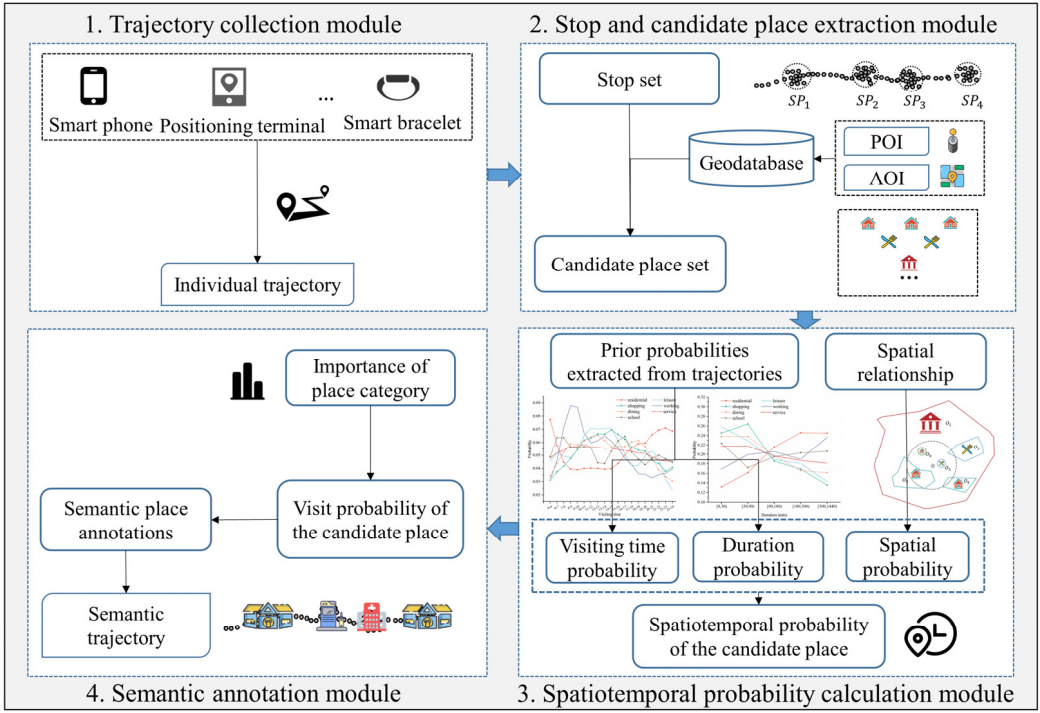

Fig. 1. Flowchart of the proposed UPAPP method.

**Definition 3**: Candidate place *O* is a place with a distance from the stop center that is less than the search radius, where the search radius is user-specified.

**Definition 4**: The spatiotemporal probability of a candidate place indicates the probability of the place being visited under the corresponding spatial location and time conditions and is only related to the place itself.

**Definition 5**: The visit probability of a candidate place represents the probability of the place being visited. This variable is calculated by combining the spatiotemporal probability of the place and the place type importance, which is related to the place category distribution in the candidate place set.

## 3.2 Extraction of the stops and candidate places

Three types of stop places were extracted [50]. The first category represents stops whereby the individual stays in a specific area without a positioning signal loss. This stop type is typically clustered based on spatiotemporal attributes. The second category denotes stops whereby the individual stays in the same area, yet the signal is partly lost due to object occlusion. The third one includes stops that belong to the start and end of the trajectory, where the individual generally leaves or arrives at a room with no signal (e.g., leaving/arriving at home).

A clustering-based algorithm was chosen to detect the first category of stops. This algorithm searches for high-density clusters with obvious aggregation in the spatiotemporal dimension. More specifically, the neighborhood of a spatiotemporal point is defined as the longest continuous subsequence starting from this point, where all points in the sequence meet the requirement that the spatial distance from the starting point is less than the specified distance threshold $d_l$. The density is defined as the time length of the sequence. Points with a density exceeding the time length threshold $t_1$ were denoted as core points, otherwise they were marked as noise. These definitions were integrated into the DBSCAN algorithm to identify the stops in the trajectory.

For the second stop category, the time interval and spatial distance of adjacent points were calculated to determine all point pairs whose time interval between adjacent points exceeds the specified time threshold $t_2$. The point pair is considered as a stop if the distance between the point pair is less than the specified distance threshold $d_2$. The third category was evaluated using the spatial distance between the end and starting points of the daily trajectory; if this distance is less than the specified distance threshold $d_3$, it is considered as a stop. Once the detection process was completed, all stops that were adjacent in time and space were merged to ensure that the stop would not be divided into several smaller stops.

After that, the attributes of each stop were calculated (e.g., stop center, radius, start time, and duration) and the corresponding geospatial information (i.e. ROI and POI) was employed to search for candidate places around each stop. These two geospatial data sources were comprehensively integrated to obtain a complete geographic database and spatial understanding.

### 3.3 Calculating the spatiotemporal probabilities of candidate places

At this stage, we now have sequences of stop episodes that have nearby candidate places associated with each stop episode. The spatiotemporal probability of a candidate place denotes the probability of a place being visited based on the conditions of the corresponding spatial location and temporal attributes. The spatiotemporal probability of candidate place $O_i$ corresponding to the stop point $SP$ can be expressed as:

$$P\left(O_i|SP\right) = P(O_i|(x,y),t,dur), \tag{1}$$

where $(x, y)$ is the coordinates of the stop center of $SP$; $dur$ is the duration of $SP$; $t$ is the visiting (start) time of $SP$;

According to the Bayesian criterion, the probability can be calculated using *Eq.* (2).

$$P(O_i|(x,y),t,dur) = \frac{P((x,y),t,dur,O_i)}{P((x,y),t,dur)}, \tag{2}$$

where $P((x,y),t,dur,O_i)$ represents the joint distribution probability of stop $SP$ and place $O_i$; and $P((x,y),t,dur)$ is the probability of the stop which is not relevant to the place, for the same stop is a constant.

Based on the Bayesian criterion, $P((x,y),t,dur,O_i)$ is calculated as:

$$P((x,y),t,dur,O_i) = P(t|(x,y),dur,O_i) \cdot P((x,y),dur,O_i)$$

$$= P(t|(x,y),dur,O_i) \cdot P(dur|O_i,(x,y)) \cdot P\left(O_i|(x,y)\right) \cdot$$

$$P\left((x,y)\right) \tag{3}$$

Substituting Eq. (3) into Eq. (2) results in:

$$P(O_i|(x,y),t,dur) = \frac{P\left(t|(x,y),dur,O_i\right) \cdot P(dur|O_i,(x,y)) \cdot P(O_i|(x,y)) \cdot P((x,y))}{P((x,y),t,dur)}. \tag{4}$$

If we assume that $(x, y)$, $t$, $dur$ are conditionally independent with respect to $O_i$, then we get:

$$P(O_i|(x,y),t,dur) = P(t|O_i) \cdot P(dur|O_i) \cdot P\left(O_i|(x,y)\right) \cdot \frac{P((x,y))}{P((x,y),t,dur)}, \tag{5}$$

where $\frac{P((x,y))}{P((x,y),t,dur)}$ is constant for the same stop. Therefore, if only the first three expressions (i.e. the visiting time, duration, and spatial probabilities, respectively) are considered, Eq. (5)

can be expressed as $P(t|O_i) \cdot P(dur|O_i) \cdot P(O_i|(x,y))$. These three expressions are calculated in turn. The above three probabilities will simultaneously affect the annotation result of the visited place. As Fig.2 shows, Place $O_1$ is most likely to be visited according to the spatial relationship only, while place $O_5$ is most likely to be visited under the corresponding temporal conditions.

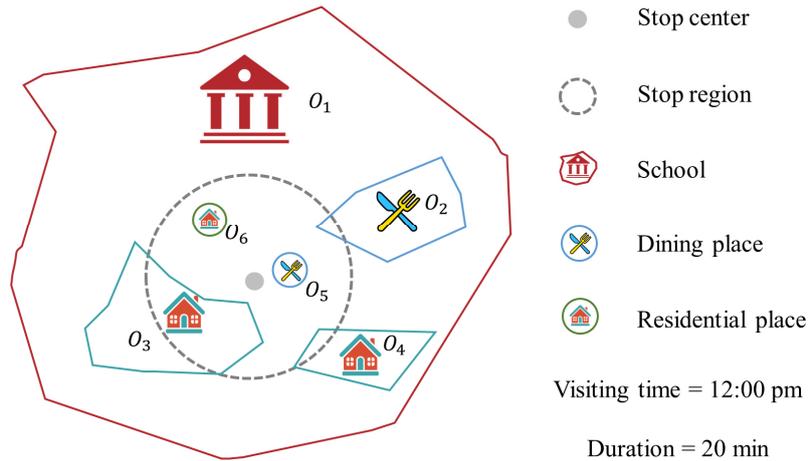

Fig. 2. Example of a stop and the corresponding candidate places

### 3.3.1 Calculation of the spatial probability

The spatial probability of a candidate place is calculated based on the relationship between the corresponding distance and topological information. Here two geographic data types are considered, the POI and ROI.

The spatial probability of ROI employs the topological relationship between the candidate geographic object and the stop region, which can be divided into the following three categories: contain, intersection and disjoint [9]. If we denote the stop region of stop $SP$ as $R_{SP}$ and the geographic range of place $O_i$ as $R_{O_i}$, then the relative spatial probability of $O_i$ can be expressed as:

$$P_{relative}\left(O_i|(x,y)\right) \;=\; \begin{cases} 1, R_{SP} \text{ contain } R_{O_i} \text{ or } R_{O_i} \text{ contain } R_{SP} \\ P_r + (1-P_r) * \frac{Area_{R_{O_i} \cap R_{SP}}}{Area_{R_{SP}}}, R_{O_i} \text{ intersect } R_{SP} \\ P_r * \frac{Searchradius - d_{R_{O_i}, center_{SP}}}{Searchradius - r_{SP}}, R_{O_i} \text{ disjoint } R_{SP} \end{cases}, \qquad (6)$$

where *contain* indicates that the topological relationship is contained, when the stop region contains the place geometry (or vice-versa); *intersect* denotes that the topological relationship is intersecting but not contained; *disjoint* indicates that the topological relationship is separated; $P_r$ is a user-specified parameter, which represents the relative spatial probability when the place geometry just intersects the stop region; $Area_{R_{O_i} \cap R_{SP}}$ represents the intersection area of $R_{O_i}$ and $R_{SP}$; $Area_{R_{SP}}$ is the area of the stop region of *SP*; $d_{R_{O_i}, center_{SP}}$ represents the minimum distance from the stop center of *SP* to geographic object $O_i$; $Searchradius$ is the user-specified radius used to search candidate places; and $r_{SP}$ is the stop radius of *SP*.

For the spatial probability of POI, it is assumed that the influence of each POI on the surrounding areas obeys a two-dimensional Gaussian distribution, where the relative probability decays with distance [4]. To ensure that the relative probability of the POI is consistent with that of the ROI, the probability is constrained to be 1 when the distance equals 0 and $P_r$ when the distance equals the stop radius. Thus, the relative probability of the POI is described as:

$$P_{relative}\left(O_i|(x,y)\right) \;=\; P\left((x,y)|O_i\right) \;=\; exp\left(-\frac{d_{O_i, center_{SP}}^2}{2\sigma^2}\right) \; s.t. \; exp\left(-\frac{r_{SP}^2}{2\sigma^2}\right) \;=\; P_r, \quad (7)$$

where $d_{O_i, center_{SP}}$ is the distance from the geographic object to the center of *SP*; and $\sigma$ is the Gaussian distribution parameter that is calculated via constraint conditions.

Following the calculation of the relative spatial probability of all candidate places, the spatial probability is then determined via normalization. For candidate place $O_i$, the spatial probability is described as:

$$P\left(O_i|(x,y)\right) \;=\; \frac{P_{relative}\left(O_i|(x,y)\right)}{\sum_i P_{relative}\left(O_i|(x,y)\right)}, \qquad (8)$$

where, $\sum_i P_{relative}\left(O_i|(x,y)\right)$ is the sum of the relative spatial probability of all the candidate places in the corresponding stop.

### 3.3.2 Calculation of the duration probability

Visits to the same place categories follow the same duration and visiting time laws [7, 51]. Assuming that the place category corresponding to $O_i$ is $C_j$, then the visiting time probability is $P(t|O_i) = P(t|C_j)$ and the duration probability is $P(dur|O_i) = P(dur|C_j)$. Previous work determines $P(t|C_j)$ and $P(dur|C_j)$ by counting labeled activity logs or extracting patterns from social media. However, this method relies on external data, which is not conducive to large-scale applications. In order to overcome the bottleneck of difficult-to-obtain annotated data, we propose a prior probability statistical method based on potential visits.

For the proposed method, a stop corresponds to a real visit, and the places in the stop region are regarded as potential visits. These potential visits are weighted according to their importance. The Term Frequency-Inverse Document Frequency index (TF-IDF) weighting method was adopted to calculate the importance weights of the potential visits [6]. If a stop region contains many places of the same type, and there are fewer places of this type in the whole area (e.g., the entire city), the importance of this type of place is high. The TF-IDF weighting method can effectively avoid the weight explosion and reflect the importance of the place category in the surrounding area. If the importance of category $C_j$ places in the stop region is denoted as $I_{C_j}^{SP}$, then category $C_j$ places are potentially visited $I_{C_j}^{SP}*1$ times in the corresponding stop.

When counting the duration prior probability, the duration is divided into multiple intervals $[dur_1, dur_2 \ldots dur_m \ldots]$. For the stop $SP$ with duration $dur \in dur_m$, the number of potential visits of stop $SP$ to category $C_j$ places is calculated as follows:

$$Visit_{C_j, dur_m}^{SP} = I_{C_j}^{SP} * 1 = \left( \frac{Count_{C_j}^{SP}}{\sum_j Count_{C_j}^{SP}} * log \frac{\sum_j Count_{C_j}}{Count_{C_j}} \right) * 1, \qquad (9)$$

where $count_{C_j}^{SP}$ is the number of category $C_j$ places in the $SP$ stop region; $\sum_j Count_{C_j}^{SP}$ is the sum of the number of all places in the $SP$ stop region; $Count_{C_j}$ is the number of category $C_j$ places in the entire area; and $\sum_j Count_{C_j}$ is the sum of the number of all places in the entire area.

The above method is employed to calculate the number of potential visits in the trajectory dataset. The average number of potential visits of all stops with duration $dur \in dur_m$ in the trajectory dataset to category $C_j$ places is then determined as:

$$Visit_{C_j,dur_m} \ = \ \frac{1}{N_{dur_m}}\sum_{i\ =\ 1}^{N_{dur_m}} Visit_{C_j,dur_m}^{SP_i},\tag{10}$$

where $Visit_{C_j,dur_m}^{SP_i}$ is the number of potential visits to type $C_j$ places in $SP_i$ (duration $dur \in dur_m$); and $N_{dur_m}$ is the number of stops with duration $dur \in dur_m$.

After separately counting the average number of potential visits to different place types belonging to different duration groups, the probability of visiting a certain place category when the duration belongs to a specified interval can be calculated. For example, the prior probability that the duration belongs to $dur_m$ when visiting a category $C_j$ place is given as:

$$P\big(dur \in dur_m|C_j\big) \ = \ \frac{Visit_{C_j,dur_m}}{\sum_m Visit_{C_j,dur_m}} \ = \ \frac{\frac{1}{N_{dur_m}}\sum_{i\ =\ 1}^{N_{dur_m}} Visit_{C_j,dur_m}^{SP_i}}{\sum_m \frac{1}{N_{dur_m}}\sum_{i\ =\ 1}^{N_{dur_m}} Visit_{C_j,dur_m}^{SP_i}},\tag{11}$$

where $Visit_{C_j,dur_m}$ is the average number of potential visits of all stops with duration $dur \in dur_m$ to category $C_j$ places; and $\sum_m Visit_{C_j,dur_m}$ is the sum of the average number of potential visits across different duration intervals to category $C_j$ places. An example of calculation the prior duration probability is presented in the Appendix A.

Assuming that place category $C_j$ corresponds to candidate place $O_i$ and the duration of the corresponding stop belongs to $dur_m$, then the duration probability of $O_i$ is $P(dur|O_i) \ = \ P\big(dur|C_j\big) \ = \ P(dur \in dur_m|C_j)$.

### 3.3.3 Calculation of the visiting time probability

Similarly, the visiting time is divided into multiple intervals $[t_1, t_2 \ldots t_k \ldots]$ and calculated with Eq. (9), Eq. (10) and Eq. (11), the prior probabilities that the visiting time belongs to different intervals when visiting a certain place type are determined. For example, for a category $C_j$ place, the prior probability of visiting time $t \in t_k$ is calculated as follows:

$$P\left(t \in t_k | C_j\right) \;=\; \frac{Visit_{C_j, t_k}}{\sum_k Visit_{C_j, t_k}} \;=\; \frac{\frac{1}{N_{t_k}} \sum_{i\,=\,1}^{N_{t_k}} Visit_{C_j, t_k}^{SP_i}}{\sum_k \frac{1}{N_{t_k}} \sum_{i\,=\,1}^{N_{t_k}} Visit_{C_j, t_k}^{SP_i}}, \tag{12}$$

where $Visit_{C_j, t_k}$ represents the average number of potential visits of all stops with visiting time $t \in t_k$ to place category $C_j$; $\sum_k Visit_{C_j, t_k}$ is the sum of the average number of potential visits of stops to place category $C_j$ with the visiting time belonging to different intervals; and $Visit_{C_j, t_k}^{SP_i}$ is the number of potential visits to type $C_j$ places in $SP_i$ with visiting time $t \in t_k$, which is the same as Eq. (9).

If place category $C_j$ corresponds to candidate place $O_i$ and the visiting time of the corresponding stop belongs to $t_k$, then the visiting time probability of $O_i$ is $P(t|O_i) = P(t|C_j) = P\left(t \in t_k | C_j\right)$.

Finally, the spatiotemporal probability of the candidate places is calculated as $P(t|O_i) \cdot P(dur|O_i) \cdot P(O_i|(x, y))$.

### 3.4 Annotation of visited places

In addition to the spatiotemporal attributes, the distribution of place categories also has an impact on the choice of visited places. For example, the probability of individuals visiting restaurants in areas containing many restaurants is higher. The probability of visiting candidate places is essentially equal to the spatiotemporal probability of the place multiplied by the importance of the category to which the place belongs. If the candidate place set of $SP$ is denoted as $O_{candidate} = (O_1, O_2 \dots O_N)$, then for category $C_j$ of place $O_i$, the importance of place $O_i$ in $O_{candidate}$ is calculated as follows:

$$I_{O_i}^{O_{candidate}} = I_{C_j}^{O_{candidate}} = \frac{Count_{C_j}^{O_{candidate}}}{\sum_j Count_{C_j}^{O_{candidate}}} * log \frac{\sum_j Count_{C_j}}{Count_{C_j}}, \qquad (13)$$

where $I_{O_i}^{O_{candidate}}$ indicates the importance of $O_i$ in the candidate place set $O_{candidate}$; $C_j$ is the place category of $O_i$; $I_{C_j}^{O_{candidate}}$ indicates the importance of place category $C_j$; $Count_{C_j}^{O_{candidate}}$ is the number of candidate places of category $C_j$ in $O_{candidate}$; and $Count_{C_j}$ is the number of places of category $C_j$ in the entire area. The probability that $O_i$ is visited in the corresponding stop is expressed as:

$$P(O_i) = P(O_i|(x,y), t, dur) * NI_{O_i}^{O_{candidate}}, \qquad (14)$$

where $NI_{O_i}^{O_{candidate}}$ is the post-normalization importance of $O_i$.

For each stop, the candidate place with the highest probability is marked as the visited place.

# 4. Experiments and results

In order to evaluate the performance of our approach, we conducted experiments based on two real GPS trajectory datasets, including the Shangdi-Qinghe dataset and the Geolife dataset. All the algorithms are implemented in Python and running on computers with Intel(R) Xeon(R) Silver 4116 CPU (2.10 GHz) and 128 GB memory.

## 4.1 Datasets

### 4.1.1 Trajectory datasets

(1) Shangdi-Qinghe trajectory datasets

The Shangdi-Qinghe trajectory dataset came from a daily behavior project conducted on the residents of Beijing City, China from October to December 2012, with a total of 709 participants living or working in the Shangdi-Qinghe area of Beijing City [52, 53]. Volunteers were required to carry GPS trackers for seven days in order to record their locations every 30 seconds with a spatial positioning accuracy of 15 m. The volunteers were from 23 communities and 19 companies, with a wide diversity in terms of gender, age and occupation. A total of 2,955,049 spatiotemporal points were collected, which is a sufficient representation of the target area.

While collecting the trajectory data, volunteers were required to record travel destination categories and activities on a daily basis to form activity logs. The collected information includes the ID, date, activity start time, activity end time, activity type, place category, etc. Place categories were grouped into residential areas, working areas, service venues, dining venues, schools, leisure venues, shopping venues, and others. The place category recorded in the activity log is used as the truth data to evaluate the accuracy of the UPAPP method in our experiment.

(2) Geolife trajectory datasets

The Geolife trajectory datasets were collected during the Microsoft Research Asia Geolife project by 182 users from April, 2007 to August, 2012 [54]. This dataset contains 17,621 trajectories with a total duration of 50,176 hours. This dataset recorded a broad range of users' outdoor movements, including life routines and some entertainment and sports activities. The majority of the data was collected in Beijing, China, and the majority of participants were college students and Microsoft employees.

Noise is present in the original trajectory data due to signal blockages, and thus pre-processing is required. Following drifting, the data points considered as noise deviate from the original position in the route, producing abnormal speeds and included angles. The included angle refers to the angle formed by the connection between the central points and two points located at the front and back. We calculated the speed and included angle of each point, and removed those points with a speed exceeding 180 km/h or an angle less than 30° to eliminate the noise in the trajectory data.

### 4.1.2 Geographic data

The POI and ROI datasets were obtained from AutoNavi and OSM (Open Street Map) data, respectively, in 2014. These geographic data were divided into the corresponding seven place categories in the activity logs, and the rules are described as follows.

The POI data already includes classification information, and we further divided it into the following seven categories with corresponding keywords: 1) residential area: "community", "residential"; 2) working area: "company", "office building", "government"; 3) service place: "life services", "medical care", "finance", "car"; 4) dining place: "dining"; 5) school: "school"; 6) leisure place: "leisure"; and 7) shopping place: "shopping".

The elements of the OSM data contain a tag expressed in the form of key = value and were also divided into the following seven categories: 1) residential area: land use = residential; 2) working area: office = *; 3) dining place: amenity = restaurant; 4) school: amenity = college, amenity = university, amenity = school, amenity = kindergarten 5) leisure place: leisure = *; 6) shopping venue: shop = *; 7) service venue: amenity = *. Places belonging to other categories were removed out.

## 4.2 Experimental setup

### 4.2.1 Compared Methods

To the best of our knowledge, the paper presents the first attempt at extracting the temporal probability from the trajectory itself. However, some existing unsupervised semantic annotation methods can be extended for place annotation. To verify the effectiveness of the proposed method, we compared the proposed UPAPP with other approaches by employing the following four methods: 1) spatial probability only; 2) spatiotemporal probability; 3) existing sequence model; 4) the proposed method combined with a sequence model. The sequence model refers to using the order (i.e. sequence) of the visited places to establish a sequence model. Spatial and sequential characteristics have recently been employed to obtain a sequence model denoted as Spatio-temporal Trajectories to Semantic Place-Matching Patterns (STOSEM) based on the HMM [9]. In this approach, the state transition probabilities were learned based on the potential visits.

The sequence models used for the comparison are detailed as follows: the observable sequence is the sequence of stop episodes; the hidden states correspond to candidate places associated with each stop episode; the emission probability employs the probability of places visited described in Section 3.4. The state transition probability describes the probability of moving from one place category to another. Based on the potential visits, we learned all the probabilities in the trajectory to generate the state transition probability matrix. In particular, for adjacent stops $SP_i$ and $SP_{i+1}$, there are a total of $m$ category $C_a$ places in the candidate places of $SP_i$, and a total of $n$ category $C_b$ places in the candidate places of $SP_{i+1}$. The potential visit method introduced in Section 3.3 was applied to calculate the number of potential visits of $SP_i$ $(SP_{i+1})$ to category $C_a$ $(C_b)$ places as $Visit_{C_a}^{SP_i}$ $(Visit_{C_b}^{SP_{i+1}})$. We assume that category $C_b$ places are visited after type $C_a$ places at a total of $Visit_{C_1}^{SP_1} * Visit_{C_2}^{SP_2}$ times, while the STOSEM method assumes this to occur $m*n$ times. All state transitions in the trajectory dataset were then counted to generate a state transition probability matrix. As a final step, the visited places were annotated via the Viterbi algorithm.

### 4.2.2 Performance Metrics

To evaluate the effectiveness of the UPAPP method in semantic place annotation, we used the Shangdi-Qinghe trajectory dataset and the corresponding activity logs for quantitative accuracy assessment. The accuracy for each place category is calculated using Eq. (15). Furthermore, overall accuracy and average accuracy were also adopted to evaluate the effectiveness of semantic annotation.

$$acc_i = \frac{TP_i}{Total_i},$$ (15)

where $acc_i$ represents the accuracy of place category $i$; $TP_i$ is the number of successful predictions of place category $i$; and $Total_i$ represents the total number of logged visits to the place category $i$.

$$OA = \frac{\sum_i^N TP_i}{\sum_i^N Total_i},$$ (16)

where $OA$ is the overall accuracy, and $N$ is the total number of place categories.

$$AA = \frac{\sum_i^N acc_i}{N},$$ (17)

where $AA$ is the average accuracy, and $N$ is the total number of place categories.

### 4.3 Results and accuracy assessment

### 4.3.1 Extraction of stops

For the extraction of the first stop category, the time and space thresholds, $t_1$ and $d_1$, were set as 600 s and 100 m, respectively. Lost points belonging to the second type of stops are formed due to signal blockage, and consequently, the positioning accuracy of nearby points is relatively poor. For this category, we set the time and space thresholds $t_2$ and $d_2$ as 1200 s and 200 m, respectively. The spatial threshold $d_3$ of the third type of stops was also set as 200 m. When merging the above stops, the time and space thresholds were set as slightly lower than those of the first stop type, namely 90 m and 540 s, respectively. Furthermore, the search radius was selected as 200 m to match stops with candidate places.

In this experiment, 12646 stops and 16926 stops were extracted from the Shangdi-Qinghe dataset and the Geolife dataset, respectively. The distributions of all the detected stops from the two trajectory dataset are presented in Fig. 3. The stops were observed to expand outwards from the center and were distributed across over half of the urban areas of Beijing city. The extracted stops in the Shangdi-Qinghe dataset were then matched with the activity log based on the corresponding temporal information. Among the extracted 12646 stops, a total of 9,283 stops matched the activity log, while 1,345 stops were unsuccessfully matched. In addition, there were 1,848 stops that had no activity log on the corresponding date and 170 stops that were not surrounded by any candidate places.

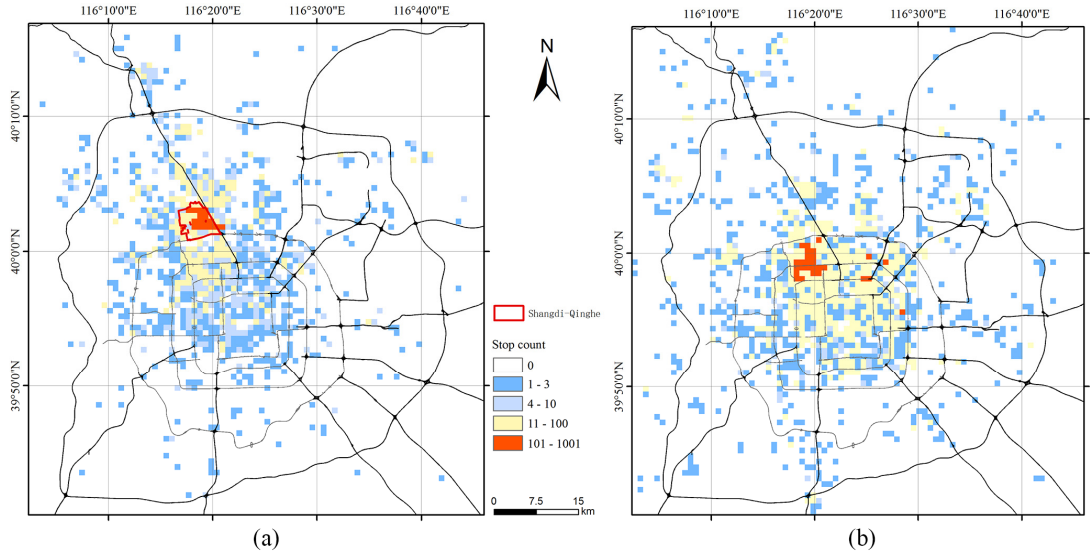

Fig. 3. Spatial distribution of extracted stops of the trajectory data. (a) the Shangdi-Qinghe dataset; (b) the Geolife Dataset

### 4.3.2 Effectiveness of extracting the visiting time and duration probability

In order to extract the prior probability of the visiting time over 24 hours, one day was divided into one group per hour. There were relatively few stops with the visiting time of 0-6:00 am, and thus they were merged into one group. As a result, the division resulted in 19 time groups. Due to the small temporal interval, a sliding window of three was employed to perform mean smoothing. Figs. 4(a) and 5(a) illustrate the prior probabilities of the visiting time for different place categories extracted from the Shangdi-Qinghe and Geolife datasets. Note that both datasets were collected in Beijing, and thus most of the extracted patterns were similar. Nevertheless, our method is able to capture differences in temporal regularity in different datasets efficiently. Significantly, the trends in the visiting time extracted from the trajectory data based on potential visits are essentially equal to residents' daily travel trends. The probability of visiting residential areas was higher in the morning and evening compared to visiting between 9:00 am to 4:00 pm. The probability of visiting working areas peaked at 8:00 - 10:00 am, with a second peak at 1:00 - 3:00 pm. Following 6:00 pm, the probabilities were lower, corresponding to the work patterns of individuals. The probability of visiting dining venues peaked at 11:00 – 1:00 pm and 6:00-7:00 pm, corresponding to lunch and dinner, respectively. Visiting shopping places was most probable in the afternoon, while the probability of visiting leisure places peaked at 2:00-4:00 pm and 8:00-10:00 pm. Noted that there were obvious differences in the trend of visiting schools in the two datasets. This can be attributed to the high participation of college students in the Geolife project, while the Shangdi-Qinghe dataset was completely based on information from individuals in employment. Furthermore, participants in the Geolife dataset had a preference of going out for leisure and shop at night.

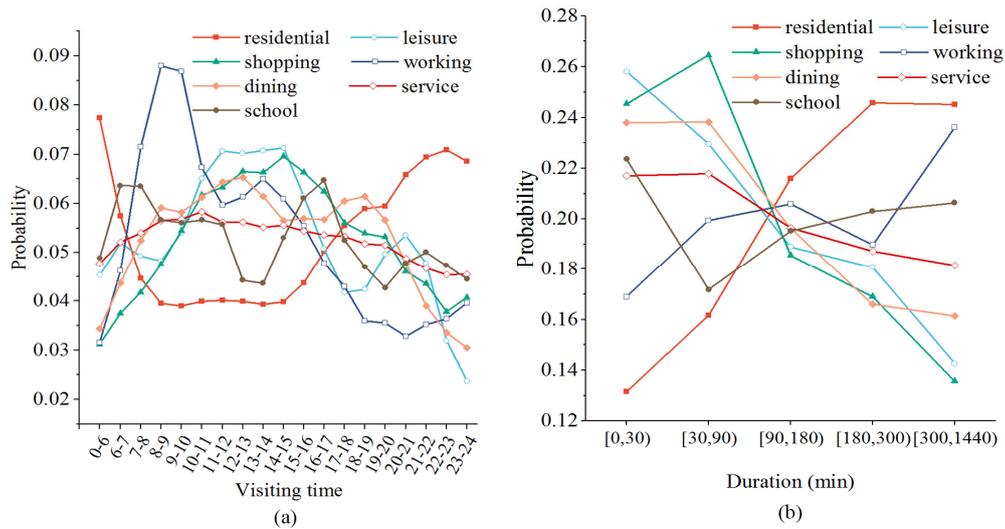

Fig. 4. Prior probabilities of visiting different place types extracted from the Shangdi-Qinghe trajectory dataset. (a) visiting time probabilities; (b) duration probabilities.

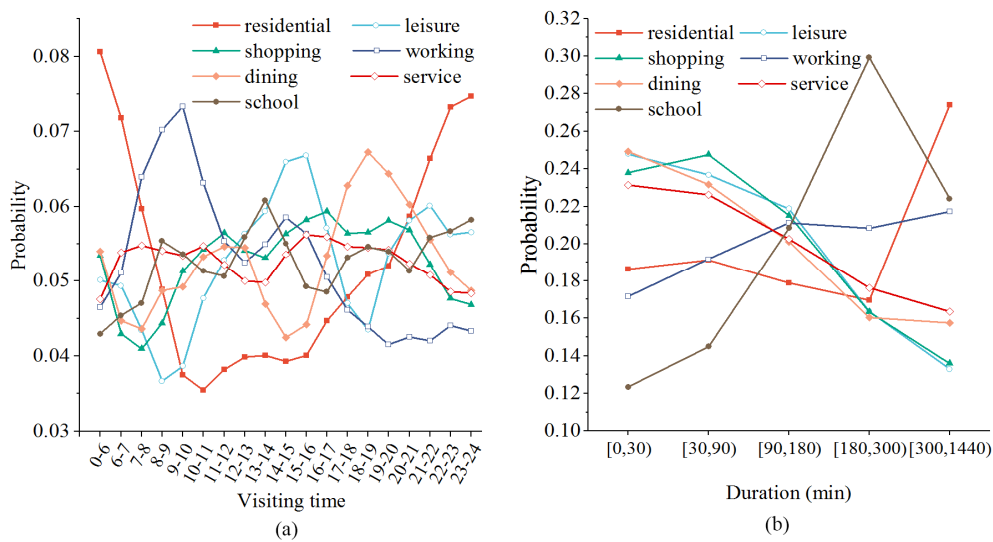

Fig. 5. Prior probabilities of visiting different place types extracted from the Geolife trajectory dataset. (a) visiting time probabilities; (b) duration probabilities.

For the purpose of extracting the prior probability of the duration, the duration was divided into the following five groups: [0-30] min, [30-90] min, [90-180] min, [180-300] min, and [300-1440] min. The prior probabilities counted based on the method detailed in Section 3.3.2 are illustrated in Figs. 4(b) and 5(b). Likewise, the patterns extracted from the two datasets were consistent with the regularity of daily activities, while differences caused by different participants were also observed. The results demonstrate that when visiting places of leisure, dining, and services, the probability decreased with the duration. When visiting residential and working areas, longer durations were associated with a greater probability. When visiting shopping venues, the probability peaked at the 30-90 min duration group, and subsequently decreased as the duration increases. For school visits, durations of 0-30 min exhibited the highest probability in the Shangdi-Qinghe dataset. This is attributed to the minimal time required to pick up children. For the remaining duration groups, the longer the visiting time at schools, the higher the probability. In contrast, since the majority of participants lived at school, durations exceeding 180 min were more probable in the Geolife dataset.

The results indicate that UPAPP can reveal key temporal patterns from trajectory data without any external data. The temporal patterns of visiting different place categories can also be quantified when activity logs are available. In order to verify the effectiveness of this method, we compared the extracted prior probability with the probability based on activity log statistics of the Shangdi-Qinghe dataset (Fig. 6). In general, the temporal patterns for visiting different place types are essentially equal to the prior probability (Fig. 4). With respect to visiting time, visits to working, residential, dining, leisure, and shopping places exhibited the similar probability peak position. The patterns in duration determined from the activity log statistics are also generally consistent. However, there were also two differences between the two sets of results, described in the following. 1) Our results revealed relatively high probabilities for durations of 300-1440 min for working place and school visits. This may be linked to the possible proximity of the school and work place to lunch venues. At close distances, the morning and afternoon may be included in the same stop, thus extending the duration. 2) For leisure visits, our results indicated high probabilities for durations of 0-30 min. This was attributed to the multiple outdoor activities in leisure venues that were always moving, which was difficult to form a consistent stop. Overall, in the absence of activity logs, our method is effective in providing temporal patterns across places.

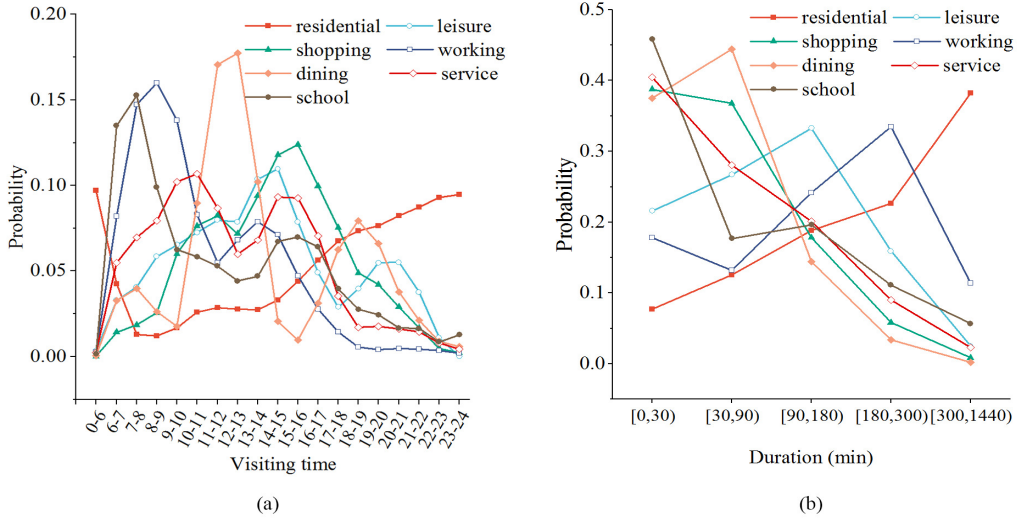

(a)                                                    (b)

Fig. 6. Temporal probabilities for different place categories extracted from activity logs

(a) Activity log-based visiting time probabilities; (b) Activity log-based duration probabilities

### 4.3.3 Effectiveness of semantic place annotation

Once the spatiotemporal probabilities of candidate places were calculated, the places visited in the stop episodes were annotated with the category importance and spatiotemporal probability. In this set of experiments, the search radius was set to 200 m and $P_r$ was set to 0.5 for all the methods otherwise specified.

To determine the constancy between the actual and annotated place categories, the semantic annotation accuracy of the 9,283 stops in the Shangdi-Qinghe dataset that successfully matched the activity log was evaluated. Table 2. shows that the UPAPP method achieves the highest overall and average accuracy. Annotating using the spatial probability-only method has the lowest accuracy. The use of visiting time probability, duration probability, and influences of the surrounding places contribute to improving the accuracy. The results reveal the ability of the proposed method to effectively annotate the visited places, with relatively high recognition accuracies for all the categories. Furthermore, the application of the ROI significantly improved the annotation accuracy of leisure venues, schools and residential areas. These place types have a relatively large geographic range, and thus integrating spatial topological characteristics can more accurately reflect the spatial characteristics of stop events. The results indicate that combining the two data types allow for the comprehensive use of the topological characteristics and category distribution, hence effectively reflecting the spatial characteristics. Besides, in the absence of labeled data for training, sequence characteristics may not improve the model accuracy but reduce efficiency.

Table 2. Performance of semantic place annotation in the Shangdi-Qinghe dataset.

| Performance metric | UPAPP | UPAPP combined with HMM | STOSEM | Spatial-only | Spatio-temporal | UPAPP with POI only |
|---|---|---|---|---|---|---|
| Dining place accuracy | **0.697** | 0.574 | 0.563 | 0.406 | 0.480 | 0.691 |
| School accuracy | **0.726** | 0.567 | 0.375 | 0.713 | 0.632 | 0.449 |
| Residential place accuracy | 0.714 | **0.788** | 0.670 | 0.619 | 0.660 | 0.599 |
| Leisure place accuracy | **0.701** | 0.581 | 0.420 | 0.660 | 0.672 | 0.201 |
| Working place accuracy | **0.682** | 0.614 | 0.642 | 0.314 | 0.637 | 0.665 |
| Shopping accuracy | 0.819 | 0.717 | **0.869** | 0.437 | 0.633 | 0.759 |
| Service place accuracy | 0.700 | 0.672 | **0.715** | 0.529 | 0.399 | 0.650 |
| **Overall accuracy** | **0.712** | 0.706 | 0.652 | 0.529 | 0.623 | 0.616 |
| **Average accuracy** | **0.720** | 0.645 | 0.608 | 0.526 | 0.588 | 0.573 |

Since the Geolife dataset lacks labeled data for validation, we conducted a qualitative analysis on the annotation results. In this dataset, the most frequently-visited schools were Tsinghua University, Beihang University, Renmin University of China, and Peking University. Working places that appeared most frequently were the Yingdu building, the Chinese Academy of Sciences, the Sigma building (used by Microsoft Research Asia), and the Shenchang building. All the above places were located in the red areas in Fig. 3 (b). This is consistent with the demographic statistics of the dataset [9, 54]. Leisure places such as the Old Summer Palace, the Summer Palace, and the Olympic Forest Park near the aforementioned places were also frequently visited. Although a quantitative evaluation is lacking, the above analysis results also demonstrate the effectiveness of UPAPP.

In the second set of experiments, we examined the impact of varying parameters in the performance of the methods. First, we have validated the effectiveness of the TF-IDF weighting algorithm in UPAPP. In our method, the TF-IDF weighting method is used to measure the importance of a place category in surroundings. We also compared the weighting method using proportional, square root with using the logarithmic function. When proportional method was directly used, the overall and average accuracy were 0.663 and 0.693, respectively. When square root weighting method was used, the overall and average accuracy were 0.704 and 0.720. respectively. These results indicate that the TF-IDF weighting method is more powerful. Using the logarithmic function can effectively avoid the weight explosion and make the weight (i.e. IDF) of a certain category close to 0 when there are many places of this category in the entire region.

We then compared the UPAPP with other methods for varying search radius and spatial probability parameter $P_r$. As shown in Fig.7, UPAPP outperforms the compared methods under different parameters, which illustrates the stability and robustness of our method. Fig. 7 (a) reveals that an inappropriate search radius affects the annotation accuracy. A small search radius may miss the correct place, while a large search radius will affect the distribution of surrounding places. Therefore, we recommend setting the search radius at 150-250 m empirically. Fig. 7 (b) indicates that the methods are not sensitive to $P_r$ when $P_r$ is less than 0.6. When $P_r$ is set too large, the spatial probability of places outside the stop radius will be correspondingly large, resulting in some mismatches. Thus, we recommend setting $P_r$ at 0.2-0.5.

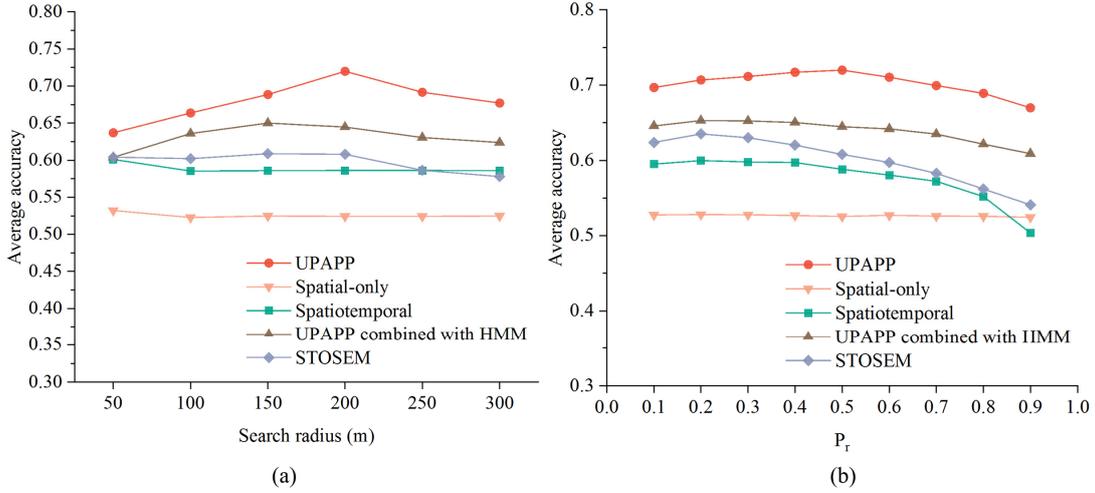

(a)

(b)

Fig. 7. Average accuracy of semantic place annotation under varying parameters on the Shangdi-Qinghe dataset. (a)search radius; (b) spatial probability parameter ($P_r$).

## 5. Discussion

### 5.1 Practical contribution

To solve the dependence on annotated and external data in calculating temporal probability, we counted the visiting time and duration probability based on potential visits. Consequently, an unsupervised approach for semantic place annotation was proposed by combining spatial information, temporal information, and place category importance. To the best of the authors' knowledge, this is the first attempt at using temporal information without any external data. Our method can be applied in different trajectory datasets that vary in regions and participants.

To further validate the effectiveness of our method, we evaluated it using the temporal probabilities derived from the activity logs. The overall accuracy of using probabilities determined by the log-based statistics is 0.725, which is slightly higher than our proposed method (0.712). This further demonstrates that the prior probabilities determined in our work can effectively reflect the temporal patterns of visits to multiple place types. Thus, without external data, our method can make full use of spatiotemporal characteristics. This will greatly increase the utility of trajectory data for large-scale applications.

## 5.2 Limitations and future work

In Section 3.3, it is assumed that the location, visiting time and duration of a stop are conditionally independent at a given place. Considering that the visiting time and duration are both temporal attributes, this assumption may not necessarily be satisfied. Thus, we discuss the rationality of this assumption. Assuming that $(x, y)$ and $t$ are conditionally independent events with respect to $O_i$, while $(x, y)$ and $dur$ are conditionally independent with respect to $O_i$, the probability in Eq. (4) can be expressed as:

$$P(O_i|(x,y), t, dur) \ = \ \frac{P(t|O_i, dur) \cdot P(dur|O_i) \cdot P(O_i|(x,y)) \cdot P((x,y))}{P((x,y), t, dur)}. \tag{18}$$

As with Eq. (12), for visited place type $C_j$ and duration $dur \in dur_m$, the probability of visiting time $t \in t_k$ can be calculated as follows:

$$P(t \in t_k | C_j, dur_m) \ = \ \frac{Visit_{C_j, dur_m, t_k}}{\sum_k Visit_{C_j, dur_m, t_k}} \ = \ \frac{\frac{1}{N_{dur_m, t_k}} \sum_{i=1}^{N_{dur_m, t_k}} Visit_{C_j, dur_m, t_k}^{SP_i}}{\sum_k \frac{1}{N_{dur_m, t_k}} \sum_{i=1}^{N_{dur_m, t_k}} Visit_{C_j, dur_m, t_k}^{SP_i}}, \tag{19}$$

where $Visit_{C_j, dur_m, t_k}$ represents the average number of potential visits of all stops with duration $dur \in dur_m$ and visiting time $t \in t_k$ to category $C_j$ places; $\sum_k Visit_{C_j, dur_m, t_k}$ is the sum of the average potential visits of stops with duration $dur \in dur_m$ to category $C_j$ places when the visiting time belongs to different intervals; and $N_{dur_m, t_k}$ is the number of stops with visiting time $t \in t_k$ and duration $dur \in dur_m$.

Assuming that conditional independence between the duration and visiting time with respect to $O_i$ are not satisfied, an overall accuracy of 0.710 is achieved, which does not reach the level of our method. However, this may be attributed to an insufficient amount of data. For the case of unconditional independence, the probability must be calculated under the two conditions, which requires a larger dataset. Increasing the data size to verify the conditional independence is reserved for future work. In addition, our method may consider the individual's historical trajectory in future work if the trajectory data is collected over a long period. For example, the individual's home and work place can be extracted using the statistics of their multi-day trajectory to improve the annotation accuracy.

## 6. Conclusions

This paper presented UPAPP, an unsupervised method for the semantic place annotation of trajectories without the requirement of supplementary data. Specifically, trajectory stops were initially retrieved from the trajectory data, and the candidate places corresponding to each stop were identified. In order to determine the visited places, a spatiotemporal probability model was created, and the Bayesian Criterion was employed to decompose it into three terms: spatial probability, duration probability, and visiting time probability. A probability calculation method that integrates both POI and ROI geographic data was developed to calculate the spatial probability, fully utilizing the characteristics of the two types of data. For the formulation of the visiting time and duration probabilities, the TF-IDF weighting algorithm was adopted to calculate the potential visits to different place types. By counting all the potential visits in the trajectory dataset, the prior probabilities of the visiting time and duration when visiting different place categories were generated. Following this, the spatiotemporal probability of the place was combined with the place category importance to annotate the visited place. Semantic annotation experiments were conducted on two real trajectory datasets. Results in the Shangdi-Qinghe trajectory dataset collected by 709 volunteers indicated that the UPAPP method achieved an overall accuracy of 0.712 and average accuracy of 0.720, which outperformed several other methods and could annotate the visited place better without relying on any other data. Moreover, the temporal patterns acquired by potential visit statistics were essentially equal to those determined directly from the activity log. Without using annotated data, our method has great potential in large-scale automated trajectory annotation.

## Acknowledgments

The authors would like to thank Professor Yanwei Chai from the School of Urban and Environmental Sciences, Peking University, for providing the trajectory data and activity logs.

This work was supported by two grants from the National Natural Science Foundation of China (No. 42171327) and the Xinjiang Production and Construction Corps, China (No.2017DB005).

35-43.

# Appendix A. An example of calculating the prior duration probability

To demonstrate the calculation details of the probabilities in our UPAPP method, we designed an example (Table 3), in which the durations are divided into five intervals: [0,30), [30-90), [90-180), [180-300), and [300,1440) min, and a total of 10 stops were used to count the prior probabilities. The potential visits to different types of places in Table 3 were calculated by Eq. (9).

Table 3. The example of calculating the duration probability

| Stop ID | Duration interval | Potential visits to schools | Potential visits to shopping places | Potential visits to dining places |
|---------|-------------------|------------------------------|--------------------------------------|------------------------------------|
| 1 | [0,30) | 0.12 | 0.19 | 0.26 |
| 2 | [0,30) | 0.14 | 0.16 | 0.31 |
| 3 | [300,1440) | 0.31 | 0.15 | 0.09 |
| 4 | [30,90) | 0.15 | 0.23 | 0.17 |
| 5 | [90,180) | 0.25 | 0.32 | 0.12 |
| 6 | [180,300) | 0.28 | 0.21 | 0.11 |
| 7 | [30,90) | 0.13 | 0.24 | 0.32 |
| 8 | [0,30) | 0.15 | 0.18 | 0.27 |
| 9 | [90,180) | 0.22 | 0.24 | 0.15 |
| 10 | [90,180) | 0.12 | 0.17 | 0.16 |

Taking visiting dining place as an example, we calculated the average number of potential visits of all stops with different duration intervals in the trajectory dataset to dining places using Eq. (10):

$$Visit_{\text{dining,dur} \in [0,30)} = \frac{1}{N_{\text{dur} \in [0,30)}} \sum_{i\,=\,1}^{N_{\text{dur} \in [0,30)}} Visit_{dining,N_{\text{dur} \in [0,30)}}^{SP_i} = \frac{(0.26 + 0.31 + 0.27)}{3} = \frac{0.84}{3},$$

$$Visit_{\text{dining,dur} \in [30,90)} = \frac{0.17 + 0.32}{2} = \frac{0.49}{2},$$

$$Visit_{\text{dining,dur} \in [90,180)} = \frac{0.12 + 0.15 + 0.16}{3} = \frac{0.43}{3},$$

$$Visit_{\text{dining,dur} \in [180,300)} = \frac{0.11}{1},$$

$$Visit_{\text{dining,dur} \in [300,1440)} = \frac{0.09}{1}.$$

Next, the prior probabilities that the duration belongs to different intervals when visiting dining places were calculated with Eq. (11):

$$P(dur \in [0,30)|dining) = \frac{Visit_{dining,dur \in [0,30)}}{\sum_m Visit_{dining,dur_m}} = \frac{\frac{0.84}{3}}{\frac{0.84}{3} + \frac{0.49}{2} + \frac{0.43}{3} + \frac{0.11}{1} + \frac{0.09}{1}} = 0.32,$$

$$P(dur \in [30,90)|dining) = \frac{\frac{0.49}{2}}{\frac{0.84}{3} + \frac{0.49}{2} + \frac{0.43}{3} + \frac{0.11}{1} + \frac{0.09}{1}} = 0.28,$$

$$P(dur \in [90,180)|dining) = \frac{\frac{0.43}{3}}{\frac{0.84}{3} + \frac{0.49}{2} + \frac{0.43}{3} + \frac{0.11}{1} + \frac{0.09}{1}} = 0.17,$$

$$P(dur \in [180,300)|dining) = \frac{\frac{0.11}{1}}{\frac{0.84}{3} + \frac{0.49}{2} + \frac{0.43}{3} + \frac{0.11}{1} + \frac{0.09}{1}} = 0.13,$$

$$P(dur \in [300,1440)|dining) = \frac{\frac{0.09}{1}}{\frac{0.57}{2} + \frac{0.49}{2} + \frac{0.43}{3} + \frac{0.23}{2} + \frac{0.09}{1}} = 0.10.$$

Likewise, the prior duration probabilities of different categories of places were calculated, and the results were shown in Table 4.

Table 4. the probabilities of visiting durations for different place categories

| Category | P(dur $\in [0,30)|C$) | P(dur $\in [30,90)|C$) | P(dur $\in [90,180)|C$) | P(dur $\in [180,300)|C$) | P(dur $\in [300,1440)|C$) |
|---|---|---|---|---|---|
| Dining place | 0.32 | 0.28 | 0.17 | 0.13 | 0.10 |
| School | 0.13 | 0.13 | 0.18 | 0.26 | 0.29 |
| Shopping place | 0.17 | 0.23 | 0.24 | 0.21 | 0.15 |